\documentclass[runningheads,a4paper]{llncs}

\usepackage{amsmath,bm}
\usepackage{multirow}
\usepackage{subfigure}
\usepackage{epsfig}
\usepackage{graphicx}
\usepackage{wrapfig}
\usepackage{amsmath}
\usepackage{algorithm}
\usepackage{algorithmic}
\usepackage{amssymb}
\usepackage{color}
\usepackage{bbm}
\usepackage{url}
\usepackage{url}

\newcommand{\keywords}[1]{\par\addvspace\baselineskip
\noindent\keywordname\enspace\ignorespaces#1}

\begin{document}
\mainmatter  
\bibliographystyle{unsrt}
\date{}

\title{ApproxDBN: Approximate Computing for Discriminative Deep Belief Networks 
}
\titlerunning{ApproxDBN: Approximate Computing for Discriminative DBNs}

\author{Xiaojing Xu%
\and Srinjoy Das\and Ken Kreutz-Delgado
}
%

\institute{Electrical and Computer Engineering, UC San Diego\\
\emph{Xiaojing Xu and Srinjoy Das have contributed equally to this work.}}

\maketitle

\vspace{-4mm}
\begin{abstract}
Probabilistic generative neural networks are useful for many applications, such as image classification, speech recognition and occlusion removal. However, the power budget for hardware implementations
of neural networks can be extremely tight. To address this 
challenge we describe a design methodology for using approximate computing methods to implement Approximate Deep Belief Networks (ApproxDBNs) by systematically exploring the use of (1) limited precision of variables; (2) criticality analysis to identify the nodes in the network which can operate with such limited precision while allowing the network
to maintain target accuracy levels; and (3) a greedy search methodology with incremental retraining to determine the optimal reduction in precision to enable maximize power savings under user-specified accuracy constraints. Experimental results show that significant bit-length reduction can be achieved by our ApproxDBN with constrained accuracy loss.
\end{abstract}

\vspace{-10mm}
\section{Introduction}
\vspace{-2mm}

Neural networks provide high performance for applications such as image recognition, text retrieval and pattern completion~\cite{hinton2006reducing}\cite{larochelle2012learning}. 
Enabling such algorithms to operate on low-power, realtime platforms such as
mobile phones and Internet of Things (IoT) devices is an area of critical interest~\cite{yasoubi2016power}\cite{chen2016eyeriss}\cite{das2015gibbs}.
Fortunately, because neural networks are inherently error-resilient~\cite{bishop1995neural}, high accuracy of arithmetic representations and operations is not necessary to generate sufficiently accurate performance of such algorithms. 
Researchers have explored the error-resilience properties of neural networks to achieve power efficiency, such as the use of limited variable precision\cite{venkataramani2014axnn}\cite{gupta2015deep}.
The majority of hardware neural network 
implementations have been done for feedforward deterministic networks~\cite{chen2016eyeriss}\cite{venkataramani2014axnn}. On the other hand, stochastic Deep Belief Networks (DBNs)  have been used for generating and recognizing images~\cite{hinton2006fast}, video sequences~\cite{sutskever2007learning}, and motion-capture data~\cite{taylor2007modeling}. Here, we develop a design methodology that uses limited variable precision to implement a Discriminative DBN which performs classification tasks.

We are motivated, in part, by the work of
Venkataramani et al.~\cite{venkataramani2014axnn} who rank neurons based on their impact on the overall performance of the network and apply limited precision to ``less critical'' (or ``resilient'') nodes, followed by a retraining process to ``heal the error'' caused by approximation so that the power benefit of approximation can be maximized with tolerable accuracy drop. In this paper, we 
carry criticality analysis and retraining to the domain of stochastic, generative neural networks. When applying approximation, two things to be decided are the number of approximated neurons and the approximation degree, which for us is the variable precision. Previous work~\cite{venkataramani2014axnn}\cite{zhang2015approxann} usually chooses a fixed number of approximated neurons, however this choice is based on some manually chosen threshold and could reduce opportunities for approximation. We propose to use an iteratively changing number of approximated neurons and an iteratively changing degree of approximation to fully exploit the potential of approximation.

In this paper, we develop a methodology for designing energy efficient implementations of a Discriminative DBN (DDBN) that meets a required level of discrimination performance. We particularly focus on the reduction of variable precision (VP), which is synonymous with variable bit-length. The problem is to find the bit-length distribution which results in lowest power within specified performance constraints. To find the distribution we propose a context dependent (i.e., application domain data-dependent) design framework for 
the principled approximation of the variables of a DDBN used for inference purposes. This is done by first performing criticality analysis on a full-precision DDBN to provide a guideline for choosing the neurons to
reduce VP and then, based on the criticality analysis, we search in a greedy way for the lowest VPs satisfying the given network
accuracy constraint. With these VPs, we retrain the DDBN to improve its performance, leaving room for further VP reduction. We repeat this approximation-retraining process until the accuracy constraint is violated with any further bit-length reduction. Summarizing, our contributions are:

\vspace{-1mm}
\begin{itemize}
\item[\small $\bullet$] The use of criticality analysis on a stochastic, generative model such as DBN to decide which neurons are less critical and are therefore more amenable to approximation.
\item[\small $\bullet$] The use of retraining on an approximated DBN to optimize network parameters with limited precision.
\item[\small $\bullet$] The development of a greedy search methodology to determine the optimal bit-length distribution indicating minimum power designs within user-specified accuracy constraints. 
\end{itemize}

\vspace{-6mm}
\section{Discriminative Deep Belief Networks}
\label{sec:background}
\vspace{-2mm}
\subsection{Restricted Boltzmann Machines}
\vspace{-1mm}
\begin{wrapfigure}{R}{0.3\textwidth}
\vspace{-14mm}
\centering
\includegraphics[width=0.3\textwidth]{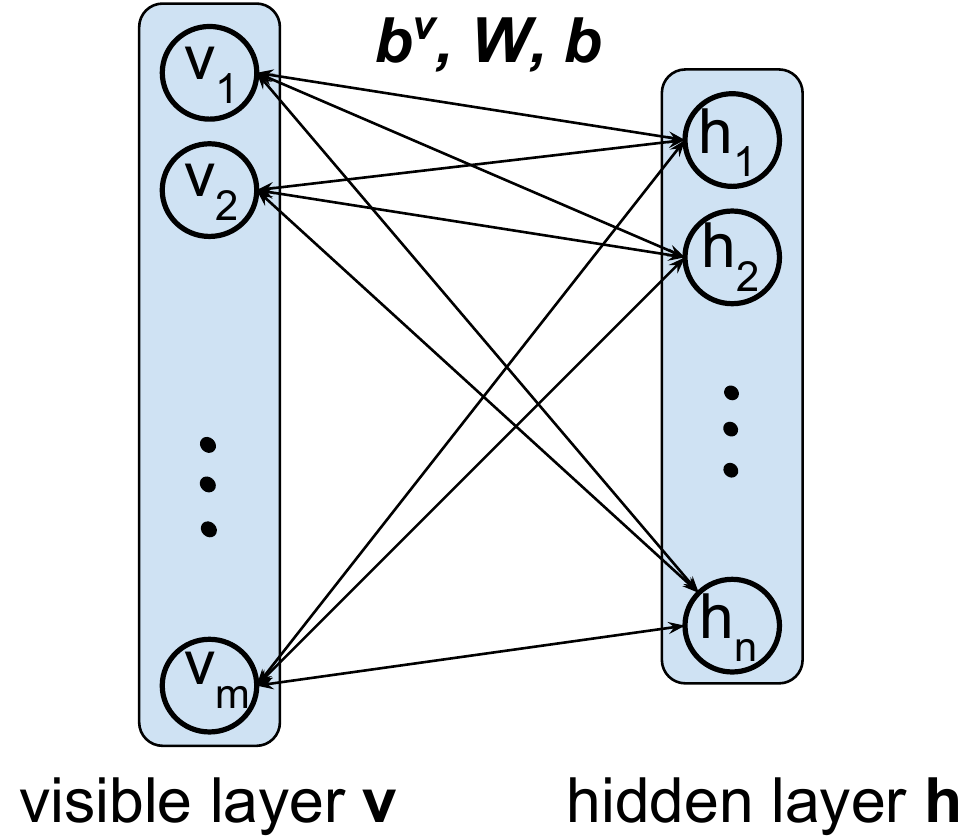}
\caption{RBM is a bipartite graph, consisting of a visible layer vector $\bm v$ and a hidden layer vector $\bm h$. $\bm b^v$ and $\bm b$ are biases for $\bm v$ and $\bm h$ respectively, and $\bm W$ is the weight between them. }
\label{RBM}
\vspace{-1mm}
\end{wrapfigure}
A Restricted Boltzmann Machine (RBM)~\cite{hopfield1982neural} is a generative stochastic neural network that can learn a probability distribution over a set of inputs (Fig.~\ref{RBM}).
The conditional distributions are~\cite{hinton2006reducing}:
\small
\begin{align}
 P(h_j|\bm v)=\sigma(b_j + \sum\limits_i v_i W_{ij}) \quad \text{and} \quad 
 P(v_i|\bm h)=\sigma(b^v_i + \sum\limits_j h_j W_{ij}) \; , \label{eq:RBMP_v}
\end{align}
\normalsize
\vspace{-1mm}
\noindent where $\sigma (x)$ is the sigmoid function.
The training method is described in~\cite{hinton2010practical}.

\newpage
\vspace{-1mm}
\subsection{Discriminative Deep Belief Networks}
\begin{figure}
\vspace{-5mm}
\centering{\includegraphics[width=20pc]{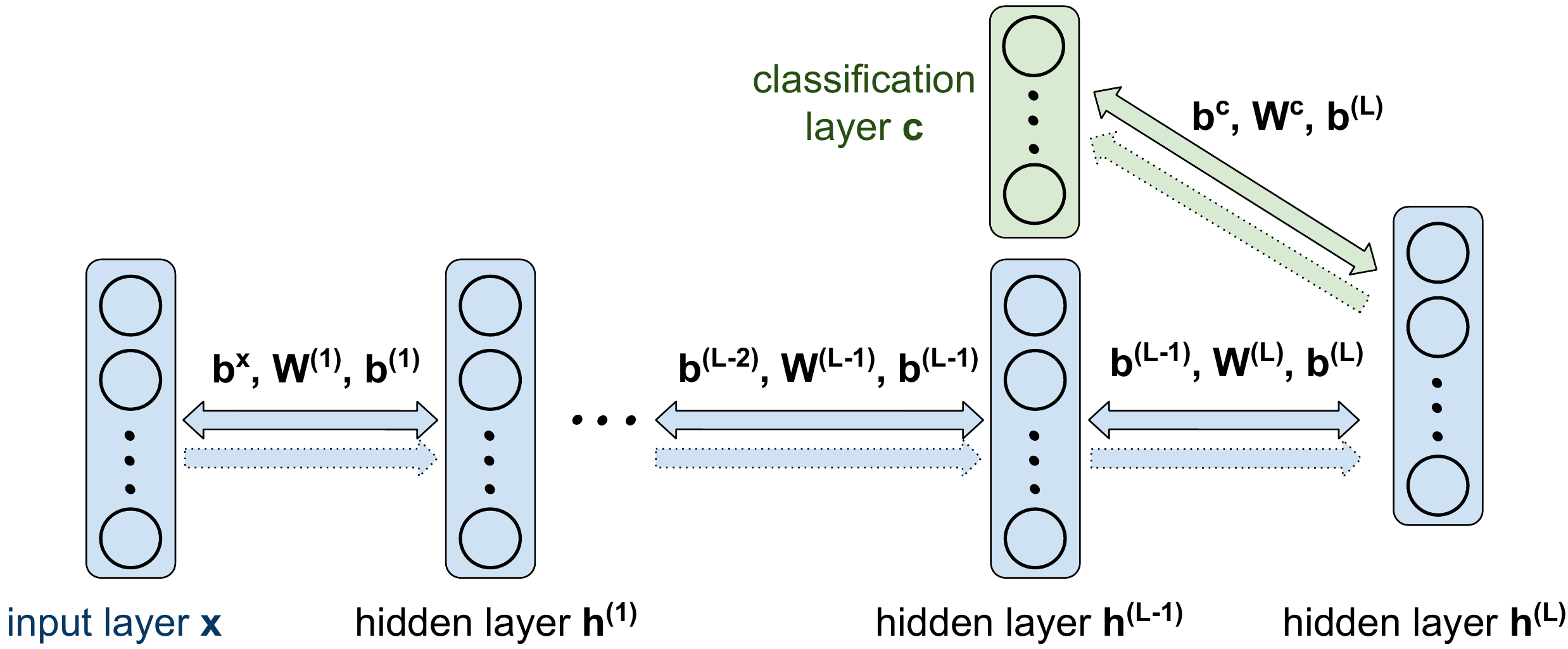}}
\caption{Discriminative DBN used for classification. It has a classification layer concatenated with the second last hidden layer to form a Discriminative RBM instead of the last RBM. The dotted arrows show the direction of signal propagation in inference, from input to the last hidden layer $\bm h^{(L)}$ to the classification layer $\bm c$}
\label{DDBN}
\vspace{-5mm}
\end{figure}

\vspace{-1mm}
Deep Belief Networks (DBNs) are probabilistic generative models which learn  a deep hierarchical representation of the training data. Reference~\cite{hinton2006reducing} shows that DBNs are just stacked RBMs and can be learned in a greedy manner by sequentially learning RBMs. 
To use a DBN as a classifier, one simple way is to replace the last RBM with a Discriminative RBM~\cite{hinton2006fast}, created by concatenating $\bm h^{(L-1)}$ and the ``one-hot'' class vector $\bm c$, as shown in Fig.~\ref{DDBN}. It is trained in the same way as a DBN without class units, but with the additional constraints on $\bm c$ such that~\cite{larochelle2012learning}
\vspace{-2mm}
\small
\begin{align}
\label{eq:DRBMP}
&P(c_i|\bm h^{(L)})=\text{softmax}(b^c_i + \sum\limits_j h_j^{(L)} W^c_{ij}) \, .
\end{align}
\normalsize

\vspace{-3mm}
\noindent with 
softmax$(x_i)= \exp(x_i)/\sum_{k} \exp(x_k)$. 
This architecture results in a Discriminative Deep Belief Network (DDBN). When used for inference, hidden units and classification units are successively sampled using previous-layer-conditional probabilities. Sampling is repeated and classification units averaged to get the final classification result.

\vspace{-6mm}
\section{Criticality Analysis}
\label{sec:criticality}
\vspace{-2mm}
Previous work implementing  approximate computing frameworks for neural networks includes ApproxANN~\cite{venkataramani2014axnn} and AxNN~\cite{zhang2015approxann} where neuron criticality analysis is implemented to identify neurons which are less critical to the overall accuracy of the network and where approximate computing elements can be instantiated. 
In these feedforward neural networks, the Euclidean distance between the output of classification and the true label is used as the loss function during training of such networks.
During criticality analysis, the derivative of this loss function is calculated with respect to the value of every neuron as that neuron's error contribution to classification error on the current sample. The magnitude of the average error contributions over all training samples is then used to characterize the criticality of the particular neuron.
In our work we are concerned with the stochastic generative model of the DDBN and only 
consider the criticality of hidden neurons because 
in a DDBN these are the major source of 
computation and power consumption.

To generalize the criticality analysis of feedforward deterministic neural network~\cite{venkataramani2014axnn}\cite{zhang2015approxann} to generative stochastic model DDBNs, there are two key steps: (i) choose a relevant loss function expressible as a function of the values of hidden neurons, and (ii) create a surrogate stochastic derivative since we don't have the derivative of a continuous and deterministic loss function as in feedforward deterministic networks. This ``derivative'' is used to determine the error contribution of the hidden neurons for the specific sample.
\begin{figure}
\vspace{-6mm}
\centering{\includegraphics[width=\textwidth]{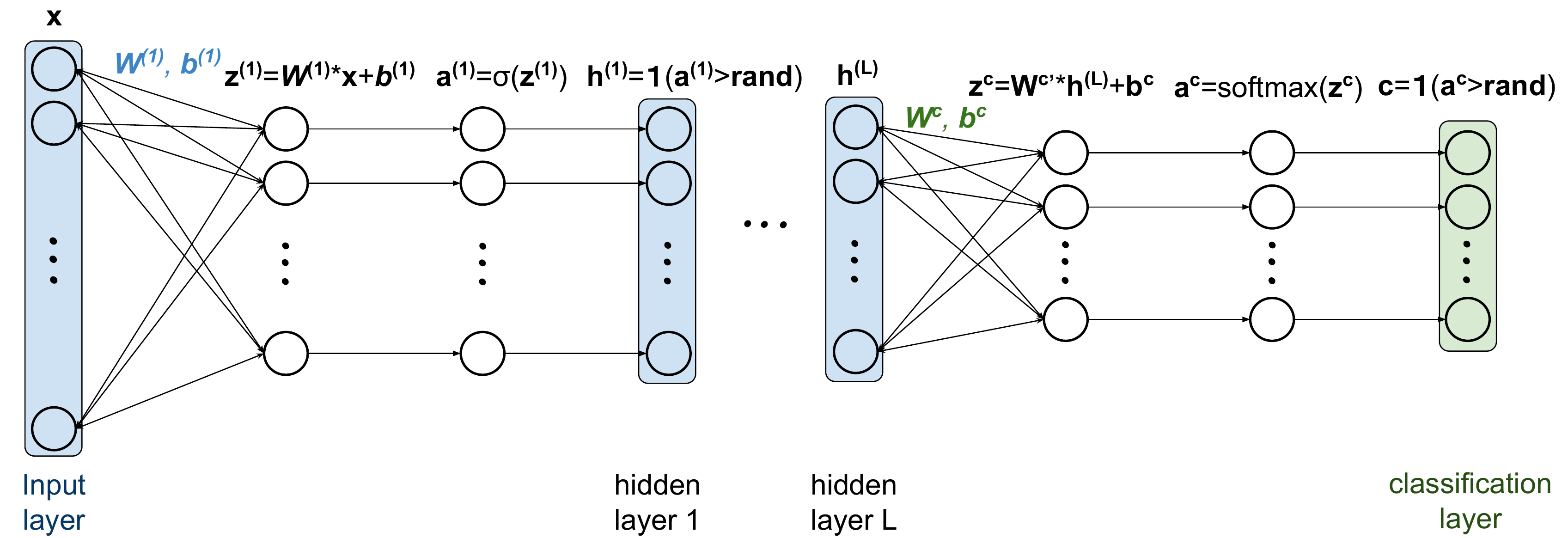}}
\caption{DDBN inference process using $L$ hidden layers}
\label{fullnet}
\vspace{-5mm}
\end{figure}

Our choice of loss function is $\text{Loss} =|\bm t-P(\bm c)|^2$, where $\bm t$ is the one-hot true label vector of the sample. It should be noted that this loss function is not the same one used in the DDBN training, but it properly interprets classification accuracy which is used as the metric evaluating the performance of Discriminative DBN inference. To calculate the derivatives, we need the values of hidden neurons to be continuous, therefore we use the probabilities $\bm a^{(l)}$ instead of binary states $\bm h^{(l)}$ in Fig.~\ref{fullnet} for the values of hidden neurons in criticality analysis. With the notations illustrated in Fig.~\ref{fullnet}, $P(\bm c)$ is the same as $\bm {a^c}$, and the derivative is
\small
\begin{equation}
\begin{aligned}
\frac{\partial\, \text{Loss}}{\partial h^{(L)}_i}
= \sum \limits _{j} \sum \limits _{l} \left( \frac{\partial \,\text{Loss}}{\partial a^c_{j}} \frac{\partial a^c_{j}}{\partial z_{l}^c} \frac{\partial z_{l}^c}{\partial h_i^{(L)}}\right)
= \sum \limits_{j} (a^c_j-t_j) \sum \limits_{l} (a^c_{l=j}-a^c_j a^c_l ) W^c_{il}  \; .
\end{aligned}
\end{equation}
\normalsize

This is the contribution of $h_i^{(L)}$ to the accuracy. To obtain the contribution of $h_i^{(l)}$ with $l<L$, we apply the chain rule:
\small
\vspace{-1mm}
\begin{equation}
\frac{\partial \, \text{Loss}}{\partial h_i^{(l)}} = \sum\limits_j\frac{\partial \, \text{Loss}}{\partial  {h_j^{(l+1)}}}\frac{\partial {h_j^{(l+1)}}}{\partial h_i^{(l)}}  \; ,
\end{equation}
\normalsize
\vspace{-1mm}
\noindent where the second factor can be derived as follows:
\small
\begin{equation}
\frac{\partial h_j^{(l+1)}}{\partial h_i^{(l)}}
=\frac{\partial \sigma(h_j^{(l)}) }{\partial h_i^{(l)}}
= \mathbbm{1}_{{i=j}}  (h_j^{(l+1)} -  (h_j^{(l+1)})^2) \; .
\end{equation}
\vspace{-1mm}
\normalsize

To determine the criticality of a hidden unit $h_i$ from the above equations, we take the magnitude of the average contribution evaluated on all training images.

\vspace{-6mm}
\section{Approximate Deep Belief Networks Design Approach}
\label{sec:designapproach}
\vspace{-2mm}
\subsection{Limited Variable Precision (VP)} \label{sec:VP}
\vspace{-1mm}

Fixed-point is preferred to floating-point representation because 
the latter requires extensive area and power for
digital hardware implementations~\cite{ly2009high}. We use $Qm.n$ to denote the 
fixed-point number format with the most significant bit being the $m$th bit to the left of the decimal point(including sign bit), and $n$ 
bits fractional part. The sum $m+n$ is the total bit-length. Our experiments show that replacing floating-point 64-bit by fixed-point 64-bit representations ($Q0.64$ for activation function outputs and $Q8.56$ for others) doesn't affect the network performance, so we use 64-bit fixed-point VP as our implementation baseline. When working on further bit-length reduction, it is usual to just change the fractional bit length in limited bit length designs and leave the integer bit length intact~\cite{yasoubi2016power}, we do the same in our limited VP design by reducing $n$ in $Qm.n$;

\vspace{-3mm}
\subsection{Accuracy Improvement with Retraining}
\vspace{-1mm}

The original DDBN is trained with floating-point 64-bit full precision and therefore its parameters such as weights and biases
may not be optimal after the approximations based on limited precision are introduced. Thus it is necessary to perform retraining of the Approximated DDBN using limited precision to reduce the impact of approximation on classification performance. The initial values of parameters in retraining are set to the previous trained network parameter values after approximation and after each round of updates in retraining, the parameters are represented by limited precision.
\begin{figure}
\vspace{-3mm}
\centering
\includegraphics[width=24pc]{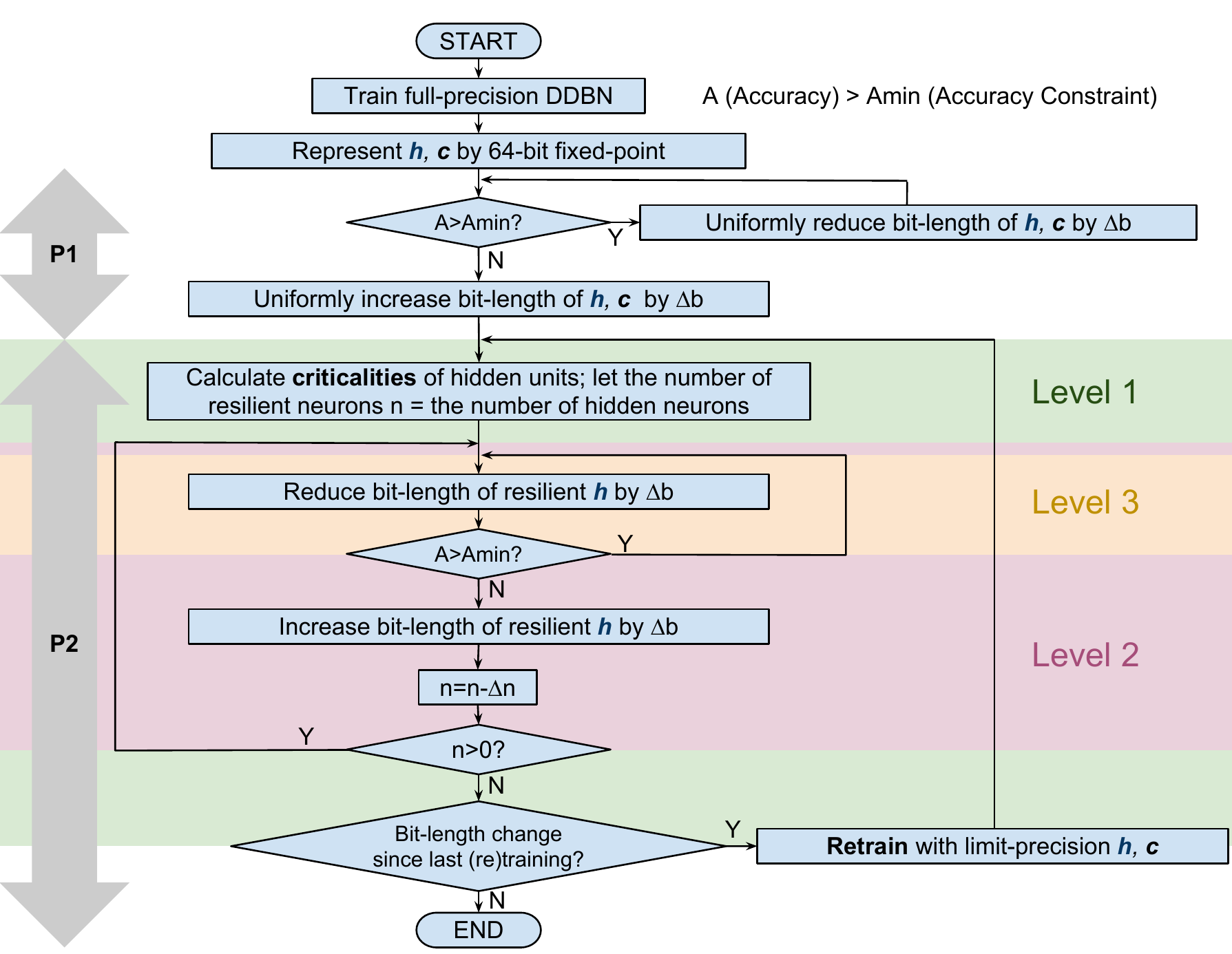}
\caption{ApproxDBN Design Approach. $h$, $c$ here stand for variables feeding into hidden and classification layers}
\label{retraining}
\vspace{-5mm}
\end{figure}
\vspace{-3mm}
\subsection{Approximate DBN Design Approach}
\label{sec:design}
\vspace{-1mm}
Our approach to designing an ApproxDBN with multiple limited VPs based on a full-precision trained DDBN is illustrated in Fig.~\ref{retraining}. It finds the optimal VP configuration under some user-specified accuracy constraints in a greedy manner. 
There are two phases of variable bit-length reduction. In the first phase (P1 in Fig.~\ref{retraining}), we apply uniform bit-length reduction for all variables until the accuracy drops below the constraint. In the second phase (P2), we try to further reduce the precisions of some variables in the hidden layers. The basic idea is to first use criticality analysis to decide the criticality order of all hidden neurons, and then reduce the precisions of the less critical neurons in a greedy manner until the accuracy constraint is violated. Then retraining is performed with limited VPs in-place to improve the accuracy. The approximation-retraining process is repeated until no bit-length reduction can be done with the accuracy constraint anymore. Specifically, maximum bit-length reduction is obtained with three loops. The outermost loop (Level 1 in Fig.~\ref{retraining}) calculates the criticality order of hidden neurons with the input DDBN, and after approximation, retrains the DDBN before the next iteration. The middle loop (Level 2) decides which neurons to apply bit-length reduction based on a iteratively changing threshold to allow less and less neurons reduce precision. The innermost loop (Level 3) is where approximation is done. It iterates to reduce bit-lengths of the neurons chosen by Level 2 until the bit-length drops to $0$ or the accuracy drops below the constraint. The program terminates when no further bit-length change occurs.

\vspace{-4mm}
\section{Experimental Results}
\label{sec:results}
\vspace{-2mm}
\subsection{Experimental Setup}
\vspace{-1mm}
We use a Discriminative DBN to perform classification on the MNIST dataset, which contains 60,000 training samples and 10,000 test samples of 28 x 28 gray-scale handwritten digits. The image data are binarized using a threshold of 0.5. Therefore the DDBN structure in our analysis is composed of 784 input units and 10 class units. Baseline DDBN has a floating-point 64-bit variable precision.

\vspace{-4mm}
\subsection{Benefits of Criticality Analysis}
\vspace{-3mm}
\begin{figure}
\vspace{-4mm}
\centering{\includegraphics[width=\textwidth]{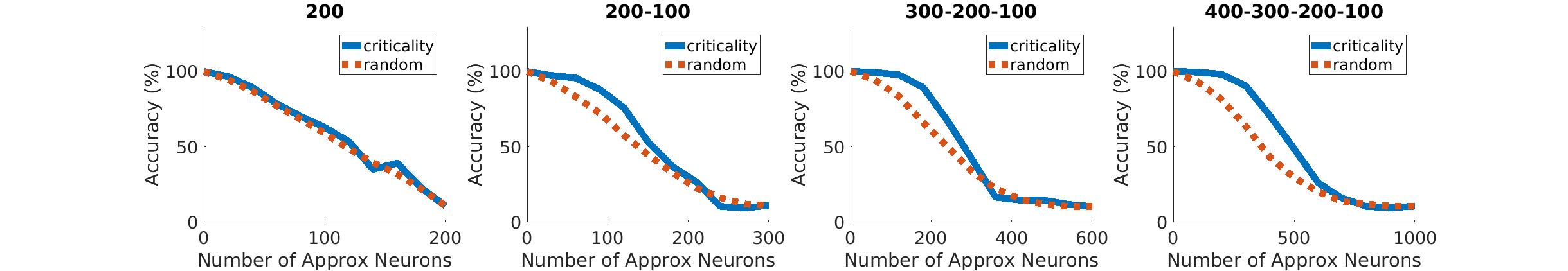}}
\caption{Accuracy relative to baseline vs. the number of approximated neurons with random approximating order and criticality analysis based approximating order for 4 different DDBN structures having one layer through four layers. $a$-$b$-$c$ means there are three hidden layers with respectively $a$, $b$ and $c$ neurons in each layer, etc. The baseline accuracies for the 4 structures are 93.41\%, 93.51\%, 94.24\% and 95.07\% respectively}
\label{criticality}
\vspace{-6mm}
\end{figure}

We show the benefits of criticality analysis by exploring four different DDBN structures with the MNIST dataset (see Fig.~\ref{criticality}). For each DDBN, we set the baseline bit-length to $Q8.8$ and approximate different numbers ($0$ to all) of hidden neurons by setting their bit-length to $0$. As seen in Fig.~\ref{criticality} approximating less critical neurons results in higher accuracies compared to approximating the same numbers of neurons randomly for DDBNs with more than 1 layer.

\vspace{-3mm}
\subsection{ApproxDBN Design Flow}
\vspace{-1mm}
\begin{figure}
\vspace{-4mm}
\centering
\includegraphics[width=28pc]{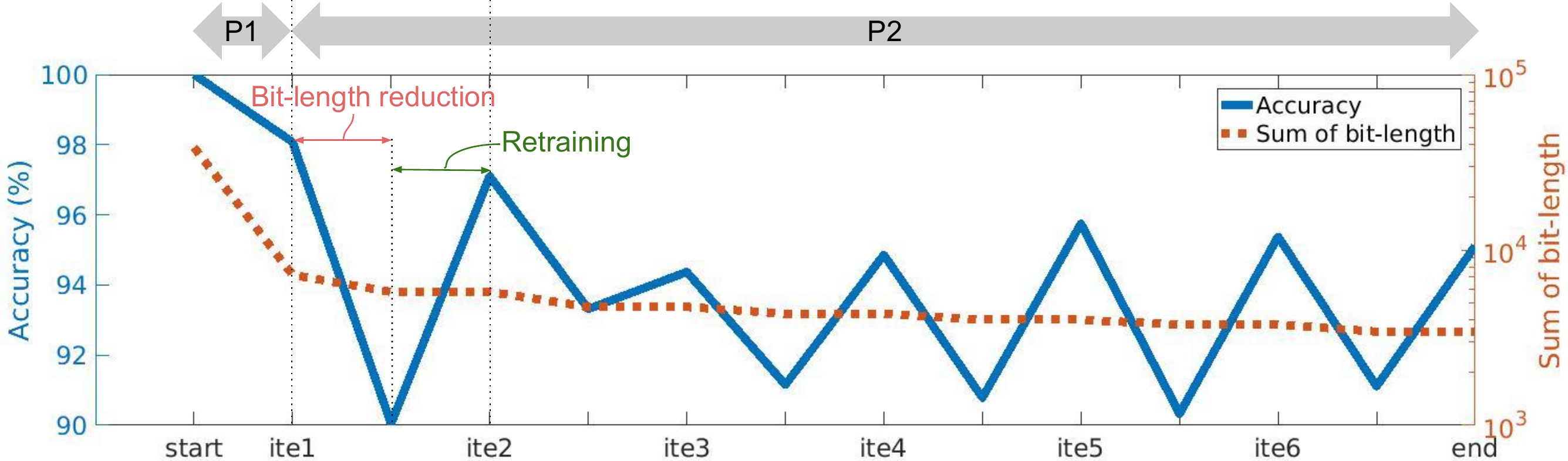}
\caption{The relative accuracy and sum of bit-length change while obtaining ApproxDBN with a 300-200-100 DDBN and a maximum accuracy loss of $10\%$.}
\label{loop}
\vspace{-4mm}
\end{figure}
We demonstrate the change in accuracy and the sum of hidden neurons bit-length during the ApproxDBN design flow as the network goes through successive iterations of criticality determination and retraining (Fig.~\ref{loop}). For each iteration in Phase 2, approximation is first applied to existing relatively high precision neurons as a result of which classification accuracy is reduced. This is followed by retraining which results in adjustment of network parameters to the updated precision and this consequently results in performance improvement. This process is repeated until at the end of the 6th iteration the accuracy can't stay above $90\%$, the target accuracy anymore with any approximation. The sum of bit-length keeps decreasing during the process, and the design methodology operates in a greedy way to make sure no variable will have bit-length increase in the process so that the overall power consumption will be reduced.

\vspace{-3mm}
\subsection{Benefits of Using Criticality Analysis and Retraining}\label{sec:results_crit}
\begin{figure}
\vspace{-2mm}
\centering
\includegraphics[width=\textwidth]{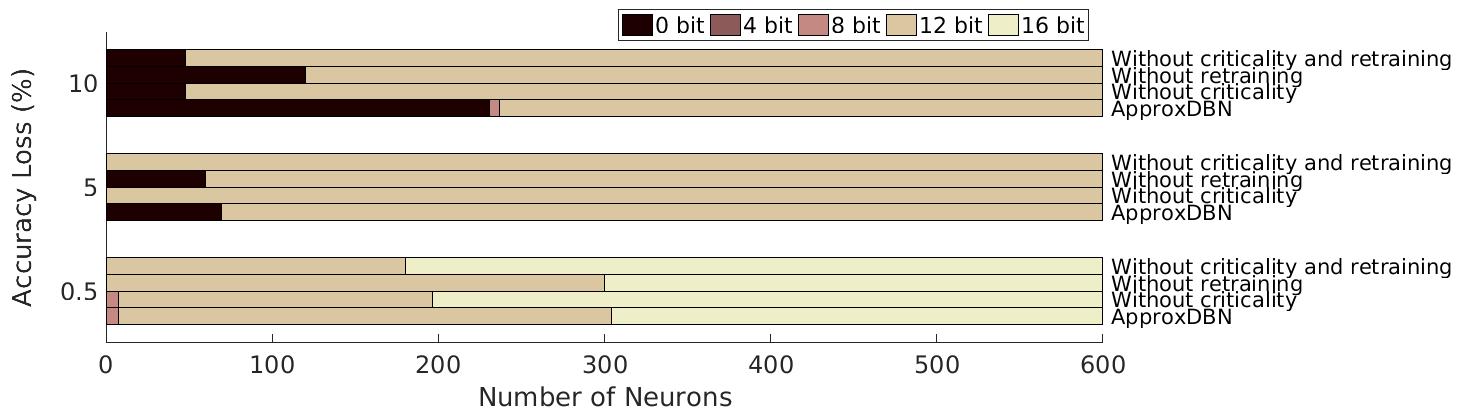}
\caption{Bit-length distribution for (1) ApproxDBN, the approximate DDBN using our design methodology, (2) ApproxDBN without criticality analysis, (3) ApproxDBN without retraining, and (4) ApproxDBN without both criticality analysis and retraining, under different relative accuracy loss constraints}
\label{bldist}
\vspace{-4mm}
\end{figure}

\vspace{-2mm}
Using a $300$-$200$-$100$ DDBN, Fig.~\ref{bldist} shows that both criticality analysis and retraining help ApproxDBN obtain a distribution of more lower bit-length variables. Criticality based selection of approximated neurons performs much better than random based selection in reduction of VPs. Retraining is important if more accuracy loss has been allowed in the bit-length reduction stages of Fig.~\ref{loop} as then there will be more room for accuracy improvement with retraining.

\vspace{-4mm}
\section{Conclusion}
\vspace{-2mm}
An approximate computing framework with constrained accuracy loss has been developed for Discriminative Deep Belief Networks (DDBNs). This framework makes use of limited variable precisions, and systematically identifies and approximates neurons that least affect network performance via application of criticality analysis and retraining.
Experimental results on DDBNs using the MNIST dataset show a significant reduction of bit-length with little accuracy loss. Future work involves extending such approximate computing techniques to more general network architectures utilizing approximate multipliers and performing generative tasks such as pattern completion and denoising.

\vspace{-3mm}

\bibliographystyle{splncs03}
\bibliography{citations}

\begin{thebibliography}{10}

\bibitem{hinton2006reducing}
G.~Hinton and R.~Salakhutdinov.
\newblock Reducing the dimensionality of data with neural networks.
\newblock {\em Science}, 313(5786):504--507, 2006.

\bibitem{larochelle2012learning}
H.~Larochelle, M.~Mandel, R.~Pascanu, and Y.~Bengio.
\newblock Learning algorithms for the classification restricted boltzmann
  machine.
\newblock {\em JMLR}, 13:643--669, 2012.

\bibitem{yasoubi2016power}
A.~Yasoubi, R.~Hojabr, and M.~Modarressi.
\newblock Power-efficient accelerator design for neural networks using
  computation reuse.
\newblock {\em IEEE Comp.~Arch.~Letters}, 2016.

\bibitem{chen2016eyeriss}
Y.-H. Chen, T.~Krishna, J.~Emer, and V.~Sze.
\newblock Eyeriss: An energy-efficient reconfigurable accelerator for deep
  convolutional neural networks.
\newblock {\em IEEE JSSC}, 2016.

\bibitem{das2015gibbs}
S.~Das, B.~Pedroni, P.~Merolla, et~al.
\newblock Gibbs sampling with low-power spiking digital neurons.
\newblock In {\em IEEE ISCAS}, pages 2704--2707, 2015.

\bibitem{bishop1995neural}
Christopher~M Bishop.
\newblock {\em Neural networks for pattern recognition}.
\newblock 1995.

\bibitem{venkataramani2014axnn}
S.~Venkataramani, A.~Ranjan, K.~Roy, and A.~Raghunathan.
\newblock Axnn: energy-efficient neuromorphic systems using approximate
  computing.
\newblock In {\em Proc.~Int.~Symposium on Low power electronics \& design},
  pages 27--32, 2014.

\bibitem{gupta2015deep}
S.~Gupta, A.~Agrawal, K.~Gopalakrishnan, and P.~Narayanan.
\newblock Deep learning with limited numerical precision.
\newblock {\em arXiv preprint arXiv:1502.02551}, 2015.

\bibitem{hinton2006fast}
G.~Hinton, S.~Osindero, and Y.-W. Teh.
\newblock A fast learning algorithm for deep belief nets.
\newblock {\em Neural computation}, 18(7):1527--1554, 2006.

\bibitem{sutskever2007learning}
I.~Sutskever and G.~Hinton.
\newblock Learning multilevel distributed representations for high-dimensional
  sequences.
\newblock In {\em AISTATS}, volume~2, pages 548--555, 2007.

\bibitem{taylor2007modeling}
G.~Taylor, G.~Hinton, and S.~Roweis.
\newblock Modeling human motion using binary latent variables.
\newblock {\em Advances in neural information processing systems}, 19:1345,
  2007.

\bibitem{zhang2015approxann}
Q.~Zhang, T.~Wang, Tian, et~al.
\newblock {ApproxANN: an approximate computing framework for artificial neural
  network}.
\newblock In {\em DATECE}, pages 701--706, 2015.

\bibitem{hopfield1982neural}
J.~Hopfield.
\newblock Neural networks and physical systems with emergent collective
  computational abilities.
\newblock {\em PNAS}, 79(8):2554--2558, 1982.

\bibitem{hinton2010practical}
G.~Hinton.
\newblock A practical guide to training restricted boltzmann machines.
\newblock {\em Momentum}, 9:926, 2010.

\bibitem{ly2009high}
D.~Ly and P.~Chow.
\newblock A high-performance fpga architecture for restricted boltzmann
  machines.
\newblock In {\em ACM/SIGDA ISFPGA}, pages 73--82, 2009.

\end{thebibliography}
\vspace{-3mm}
\end{document}